\documentclass[]{typhoon} % [twocolumn] for two columns

\usepackage{amsmath,amssymb}
\usepackage[toc,page,header]{appendix}
\usepackage{minitoc}
\usepackage{tikz}
\usetikzlibrary{patterns}
\usepackage{float}
\usepackage{bm}
\usepackage{soul}
\usepackage{algorithm}
\usepackage{algpseudocode}
\usepackage{cleveref}
\usepackage{enumitem}
\usepackage{xspace}

\newcommand{\huggingface}{\raisebox{-1.5pt}{\includegraphics[height=1.05em]{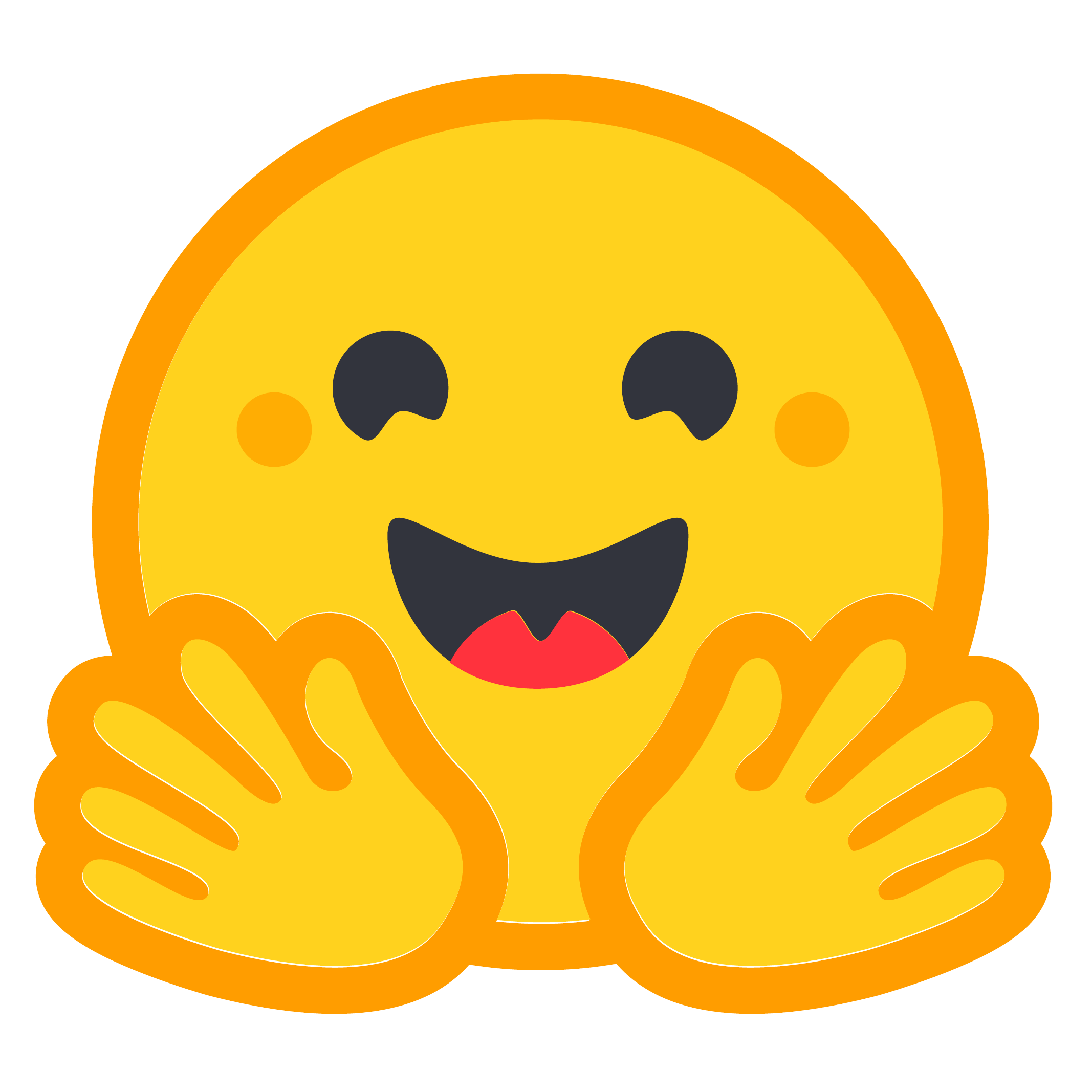}}\xspace}
\newcommand{\github}{\raisebox{-1.5pt}{\includegraphics[height=1.05em]{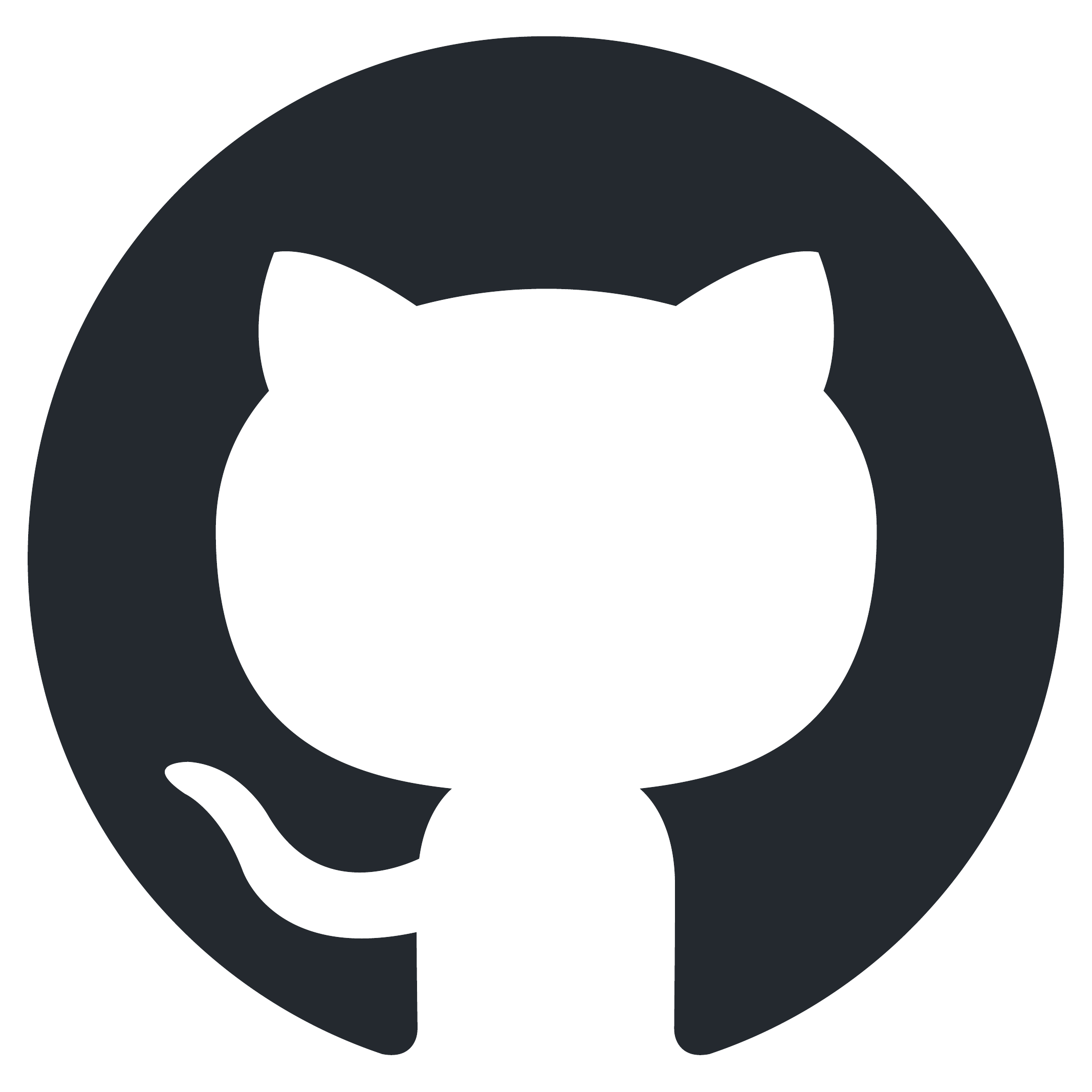}}\xspace}

\title{Typhoon-S: Minimal Open Post-Training for Sovereign Large Language Models}
\author{Kunat Pipatanakul, Pittawat Taveekitworachai}
\affiliation{Typhoon, SCB 10X}
\abstract{Large language models (LLMs) have progressed rapidly; however, most state-of-the-art models are trained and evaluated primarily in high-resource languages such as English and Chinese. In addition, they are often developed by a small number of organizations with access to large-scale compute and data. This gatekeeping creates a practical barrier for \emph{sovereign settings} in which a regional- or national-scale institution or domain owner must retain control and understanding of model weights, training data, and deployment while operating under limited resources and strict transparency constraints. To this end, we identify two core requirements: (1) \emph{adoptability}, the ability to transform a base model into a general-purpose assistant, and (2) \emph{sovereign capability}, the ability to perform high-stakes, region-specific tasks (e.g., legal reasoning in local languages and cultural knowledge). We investigate whether these requirements can be achieved without scaling massive general-purpose instruction corpora or relying on complex preference tuning pipelines and large-scale reinforcement fine-tuning (RFT). We present \textbf{Typhoon S}, a minimal and open post-training recipe that combines supervised fine-tuning, on-policy distillation, and small-scale RFT stages. Using Thai as a representative case study, we demonstrate that our approach successfully addresses adoptability by transforming both sovereign-adapted and general-purpose \emph{base} models into \emph{instruction-tuned} models with strong general performance. We further show that small-scale RFT with \textbf{InK-GRPO}--an extension of GRPO that augments the GRPO loss with a next-word prediction loss--enables sovereign capability by improving Thai legal reasoning and Thai-specific knowledge while preserving general capabilities. Our results suggest that a carefully designed post-training strategy can reduce the required scale of instruction data and computation, providing a practical path toward high-quality sovereign LLMs under academic-scale resources (approximately two days of 8-GPU training for an 8B model for adoptability, and one day of 4-GPU training for sovereign capability).}

\checkdata[Project Resources]{\newline
\huggingface\texttt{Collections}: \href{https://huggingface.co/collections/typhoon-ai/typhoon-s}{\texttt{Typhoon-S-Collections}}
\newline
\huggingface\texttt{Models}: \href{https://huggingface.co/typhoon-ai/typhoon-s-thaillm-8b-instruct-research-preview}{\texttt{Typhoon-S-8B-Instruct}},
\href{https://huggingface.co/typhoon-ai/typhoon-s-4b-nitibench-ccl-legal-agent-research-preview}{\texttt{Typhoon-S-4B-Legal-Agent}}
\newline
\huggingface\texttt{Datasets}: \href{https://huggingface.co/datasets/typhoon-ai/typhoon-s-instruct-post-training}{\texttt{Typhoon-S-Instruct-Dataset}}, \href{https://huggingface.co/datasets/typhoon-ai/typhoon-s-sovereign-capability-dataset}{\texttt{Typhoon-S-Sovereign-Capability-Dataset}}
\newline
\github\texttt{Github}: \url{https://github.com/scb-10x/typhoon-s}
}

\begin{document}
\maketitle
\vspace{-20pt}

\section{Introduction}
Recent frontier large language models (LLMs) can be broadly categorized into three types: (1) proprietary systems, (2) open-weight models, and (3) fully open initiatives. While proprietary models have historically defined the state-of-the-art, recent open-weight models such as Qwen \citep{yang2025qwen3technicalreport}, Gemma \citep{gemmateam2025gemma3technicalreport}, and DeepSeek \citep{deepseekai2025deepseekv3technicalreport} have made rapid progress, delivering performance competitive with top-tier closed systems. Simultaneously, fully open efforts like OLMo \citep{olmo2025olmo3}, Apertus \citep{swissai2025apertus}, and Nemotron \citep{bercovich2025llamanemotronefficientreasoningmodels} aim to democratize scientific understanding by releasing not just weights, but also training pipelines and datasets.

However, the development of state-of-the-art LLMs remains concentrated in a small number of organizations, with most models trained primarily on English- and Chinese-centric data or supported by large-scale compute resources and complex post-training pipelines. Even ``fully open'' models typically rely on massive-scale compute (e.g., OLMo 3 used a cluster of 1,024 H100 GPUs for several months, totaling \$2.75M \citep{olmo2025olmo3}) and complex post-training pipelines that are inaccessible to smaller research groups or national initiatives. This creates a form of ``\textit{resource gatekeeping},'' where the standard recipe for achieving high performance requires scale that exceeds the capacity of most academic or public sector entities. This concentration presents significant challenges for \emph{sovereign setting}. 

We define a \textbf{sovereign setting} as a scenario in which developers must retain control over and understanding of model weights, data, and training methodologies, while ensuring strong alignment with specific \textbf{regional, cultural, and in-domain requirements}. Unlike general commercial development's ``one-size-fits-all'' models, sovereign adaptation often operates under limited resources, constrained by both compute budgets and human expertise.

As a result, smaller research groups and national-level initiatives--especially in resource-constrained countries--face difficulties in adapting these models to regional- or domain-specific needs. In practice, \emph{sovereign-adapted base models} often demonstrate strong regional knowledge (e.g., strong Thai performance on MMLU-style benchmarks such as ThaiExam  \citep{pipatanakul2023typhoonthailargelanguage} or the M3/M6 exams  \citep{yuenyong2025openthaigpt16r1thaicentric}), but they lag behind leading proprietary and open-weight models in general instruction following, tool use, and agentic behaviors, limiting their practical adoption.

A common approach has been to \emph{scale} general-purpose instruction data and post-training pipelines. Contemporary post-training frameworks typically combine supervised fine-tuning (SFT), preference optimization (e.g., DPO), and reinforcement fine-tuning (RFT), supported by large, curated instruction datasets. While effective, this strategy poses two major challenges in sovereign settings. First, scaling general-purpose instruction data tends to prioritize high-resource languages and generic tasks, increasing dependence on external data quality and availability. Second, the engineering complexity and data requirements of these pipelines can exceed the capacity of resource-constrained teams. As a result, there remains no clear, validated post-training recipe tailored to sovereign deployment under limited-resource conditions.

To this end, we propose \textbf{Typhoon S}, a minimal and open post-training recipe addressing the aforementioned challenges. Specifically, Typhoon S is designed to address this central research question: \textit{Which post-training strategy can enable competitive performance with frontier models under academic-level resource constraints?}

We breakdown this research question into two aspects: \textbf{1) Adoptability}, defined as the ability to transform a base model into a general-purpose assistant capable of instruction following, mathematical, code generation, and tool use with a competitive performance; and \textbf{2) Sovereign Capability}, defined as an ability to perform high-stakes, region-specific tasks (e.g., local legal reasoning, cultural knowledge, and language-specific logic) that are often under-represented in general pretraining data. We use Thai as a representative sovereign setting. In this technical report, we study the two aforementioned core capabilities into separate experimental setups to reduce cross-contamination effects.

For \emph{adoptability}, we demonstrate that a multi-stage post-training framework--consisting of lightweight SFT on general instructions and on-policy distillation (OPD) from teacher models--can effectively transform sovereign-adapted base models into competitive instruction-tuned assistants. 

For \emph{sovereign capability}, we introduce \textbf{In}jected \textbf{K}nowledge GRPO (InK-GRPO), an extension of GRPO that augments the GRPO loss with a cross-entropy (next-token prediction) loss. We show that InK-GRPO effectively improves target performance while teaching new knowledge in parallel. We apply this technique in both standard RFT settings and agentic RFT setups; in the latter, the model has access to tools for retrieving external knowledge during both training and inference. We observe particularly strong improvements in Thai legal reasoning scenarios.

We summarize the contributions of our paper as follows:
\begin{enumerate}
    \item We identify \textbf{two complementary requirements} for sovereign post-training: \textit{adoptability} as a general-purpose assistant and \textit{sovereign capability} on regional- or cultural-specific tasks.
    \item To address adoptability, we present a minimal \textbf{base-to-instruct recipe} that combines lightweight SFT with OPD, using open-source instruction data together with target-language data.
    \item To enhance sovereign capability, we introduce \textbf{InK-GRPO}, which optimizes domain-specific performance while teaching new knowledge in parallel within an agentic RFT setup on reasoning and knowledge benchmarks.
\end{enumerate}

\section{Adoptability}
\label{sec:adopt}

In this section, we assume a scenario in which a national-level initiative needs to develop a Thai instruction-tuned model from a base model to enable instruction-following capabilities under a limited budget--for example, using less than one 8 $\times$H100 for under a week wall-clock time. Our approach is a two-stage post-training pipeline consisting of (1) SFT followed by (2) OPD. For training datasets, we uses a standard monolingual English dataset as the foundation, augmented with a small amount of target-language data to enable instruction-following ability in the target language \citep{pipatanakul2024typhoon2familyopen, taveekitworachai2025typhoont1openthai, pipatanakul2025adaptinglanguagespecificllmsreasoning,tjhi-etal-2023-sea}. Given the assumed scenario, a key data constraint is the exclusive use of \textit{open-source} English datasets, reflecting a practical assumption for sovereign developers: to leverage existing resources rather than reinventing the wheel. Specifically, we use the Tulu 3 SFT dataset  \citep{lambert2025tulu3pushingfrontiers} as the primary instruction corpus and Toucan  \citep{xu2025toucansynthesizing15mtoolagentic} as the tool-use dataset due to it openness, diversity and credibility.
For augmented small target-language datasets used to improve language-specific benchmarks, we follow a well-established approach \citep{pipatanakul2024typhoon2familyopen}. The remainder of this section details target-language data construction, SFT, OPD, training and evaluation procedures, as well as results and discussion.

\subsection{Target Language Dataset}
\label{sec:post_languagel_alignment}

To support Thai instruction following, we construct a dedicated target-language dataset following our previous work  \citep{pipatanakul2024typhoon2familyopen}, as shown in \Cref{fig:target_language_dataset}, by performing the following steps:

\begin{figure}[ht]
    \centering
    \includegraphics[width=\textwidth]{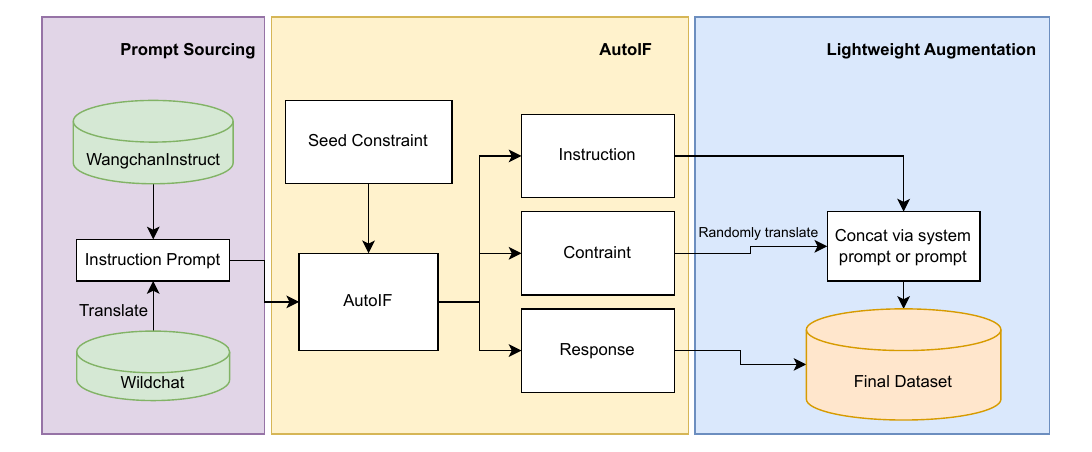}
    \caption{Overview of the target-language dataset construction pipeline for Thai.}
    \label{fig:target_language_dataset}
\end{figure}

\paragraph{Prompt sourcing}
We draw prompts from two sources: 
(1) We use a Thai translated \textbf{real-user} English prompts from WildChat  \citep{zhao2024wildchat1mchatgptinteraction} due to the lack of publicly available Thai real-user prompt data.
(2) We aggregate prompts from available Thai instruction datasets, including WangchanThaiInstruct  \citep{limkonchotiwat2025wangchanthaiinstructinstructionfollowingdatasetcultureaware}, Han \citep{phatthiyaphaibun_2024_13145164}, and Typhoon Instruct \citep{pipatanakul2023typhoonthailargelanguage}.

\paragraph{Response generation and filtering}
Once we have Thai prompts, we generate responses from those prompts using the AutoIF framework  \citep{dong2024selfplayexecutionfeedbackimproving}. We define code-verifiable constraints in both English and Thai using a small set of manually curated few-shot examples in Thai and English; these constraints are combined with the sourced prompts to guide response generation. Candidate responses are produced using Qwen3~235B~A22B~Instruct~(2507). Quality control is enforced through rejection sampling based on AutoIF test cases and model self-evaluation. Responses with scores below 7 are discarded.

\paragraph{Lightweight augmentation}
Prompts in the AutoIF format consist of two components: \textbf{(1) the main instruction} (the user query) and \textbf{(2) constraints}, which specify rules and requirements for the response. Constraints can be placed either in the system prompt or included in the user prompt together with the main instruction. In monolingual settings, all prompt components typically use the same language. In multilingual settings, however, prompts often involve code mixing. For example, a user may provide the main instruction in Thai while specifying constraints in English. Moreover, multilingual LLMs tend to perform internal reasoning in English regardless of the prompt language \citep{schut2025multilingualllmsthinkenglish}.

To mitigate these issues and encourage cross-lingual alignment, as well as robustness to constraint placement (system vs.\ user prompts), we use two simple augmentation approaches:
\begin{enumerate}
\item Randomly translating constraints between English and Thai.
\item Randomly assigning constraints either to the system message or concatenating them with the user instruction.
\end{enumerate}

From our pilot experiments, we find that mixing English constraints with Thai user instructions yields performance improvements across tasks--including mathematics, code generation, and agentic behaviors--while preserving data efficiency. Therefore, we construct the dataset using this approach.

\subsection{SFT}
\label{sec:sft}

SFT is selected as the first stage in the training pipeline to equip the base model with basic instruction-following and tool-use abilities. We follow a standard SFT setup, optimizing a cross-entropy objective over instruction–response pairs.

\paragraph{SFT Dataset}
We construct a unified SFT corpus by combining datasets corresponding to three complementary objectives:
\begin{itemize}
\item \textbf{English-centric general instruction following:} We sample 200k instances from the Tulu~3 SFT dataset  \citep{lambert2025tulu3pushingfrontiers}, which contains over one million examples spanning diverse task domains.
\item \textbf{English-centric tool use:} We incorporate the Toucan Tool dataset  \citep{xu2025toucansynthesizing15mtoolagentic}, which provides supervision for structured tool usage.
\item \textbf{Thai language instruction:} We include a Thai instruction dataset (\Cref{sec:post_languagel_alignment}) to improve Thai instruction-following quality beyond English-only supervision.
\end{itemize}

The final data mixture is summarized in \Cref{tab:sft_data}. This design targets a balanced skill profile--general instruction following, tool use, and Thai-language capability--while keeping the pipeline simple and minimizing computational requirements.

\begin{table}[ht]
\centering
\renewcommand*{\arraystretch}{1.2}
\small
\begin{tabular}{llr}
\hline
\textbf{Source Dataset} & \textbf{Description} & \textbf{\#Samples} \\
\hline
Tulu 3 SFT & General instruction tuning across diverse tasks & 200k \\
Toucan Tool & Tool-use and agentic interaction examples & 100k \\
Typhoon AutoIF & Thai language alignment (AutoIF-style) & 40k \\
\hline
\textbf{Total} & & \textbf{340k} \\
\hline
\end{tabular}
\caption{Summary of data sources and statistics for the SFT dataset.}
\label{tab:sft_data}
\end{table}

\subsection{On-Policy Distillation}
\label{sec:onpolicy}

Following SFT, we apply OPD using the Generalized Knowledge Distillation (GKD)  \citep{agarwal2024onpolicydistillationlanguagemodels,yang2025qwen3technicalreport,lu2025onpolicydistillation}. The motivation is to reduce the train-inference distribution mismatch of standard (offline) distillation: if the student is only trained on fixed teacher outputs, it may not learn how to recover from its own mistakes at inference time. In GKD, the teacher instead provides token-level feedback on \emph{student-generated} trajectories, so the corrective signal better matches the student's current behavior.

Concretely, GKD alternates between two data sources using a \emph{student-data fraction} $\lambda \in [0,1]$. At each training step, with probability $\lambda$ we (i) sample an input $x$ and generate an output $y \sim p_S(\cdot \mid x)$ from the student (on-policy data); otherwise, we (ii) sample $(x,y)$ from a reference dataset (e.g., SFT data). We then query the teacher for token probabilities along the chosen sequence and update the student to minimize a divergence $D$ between teacher and student token-level distributions, averaged over time steps.

OPD can be implemented in two variants: \emph{full-distribution} (full-logits) distillation, which uses the teacher's complete next-token distribution, and \emph{top-$K$} distillation, a computationally efficient approximation that retains only the teacher’s top-$K$ token probabilities. In our experiments, we use full-logits distillation, unless state otherwise. We additionally compare both variants to study the trade-offs between performance and efficiency, with detailed results presented in \Cref{sec:full_vs_topk_opd}.

\subsubsection{Distillation Pipeline}
\label{sec:distill_pipeline}

In common RL training frameworks for LLMs (e.g., veRL \citep{sheng2024hybridflow} and OpenRLHF  \citep{hu2025openrlhfeasytousescalablehighperformance}, the teacher model is often treated as a remote worker, which introduces substantial overhead due to the serialization and transfer of logits across Ray processes. This communication cost can dominate training latency, in some cases exceeding the time required for a forward pass and response generation during on-policy rollouts. In our single-node setup, these overheads provide limited benefit and instead become a bottleneck. To address this issue and enable efficient full-logits OPD, we implement a lightweight training framework based on the HuggingFace Transformers and TRL \citep{vonwerra2022trl} stack. Drawing inspiration from the hybrid flow model proposed in veRL  \citep{sheng2024hybridflow}, we integrate teacher forward passes directly into the training loop while preserving memory efficiency. Specifically, our training framework implements \textbf{dynamic model swapping}, in which inactive models are offloaded to RAM when not in use, reducing GPU VRAM pressure without sacrificing model availability. This design enables the orchestration of large teacher models under limited hardware budgets.

In addition, our pipeline incorporates:
\begin{itemize}
    \item \textbf{FSDP with CPU offloading} to enable full fine-tuning of an 8B-parameter student model under limited GPU memory.
    \item \textbf{vLLM as the inference backend} for high-throughput, low-latency student inference within the training loop.
    \item \textbf{Hybrid scheduling} to coordinate model swapping scheduling, only working model is loaded into GPU VRAM, to optimize the throughput and offload to RAM after used.
\end{itemize}

Using this pipeline, we train the 8B model on 4 $\times$H100 GPUs, demonstrating that teacher-student full-logits OPD can be executed efficiently without large-scale distributed computing resources.

\subsubsection{OPD Dataset}
\label{sec:distill_data}

For the OPD dataset, we subsample from the SFT mixture to construct a smaller corpus. We downsample the English subset while retaining the full Thai dataset. The final OPD dataset contains 160k examples with the mixture shown in \Cref{tab:distill_data}.

\begin{table}[ht]
\centering
\small
\renewcommand*{\arraystretch}{1.2}
\begin{tabular}{llr}
\hline
\textbf{Source Dataset} & \textbf{Task Type} & \textbf{\#Samples} \\
\hline
Tulu 3 (subset) & General instruction-following & 100k \\
Typhoon AutoIF & Thai language alignment (AutoIF) & 40k \\
Toucan Tool (subset) & Agentic \& tool-use tasks & 20k \\
\hline
\textbf{Total} & & \textbf{160k} \\
\hline
\end{tabular}
\caption{Summary of data sources and statistics for the OPD dataset.}
\label{tab:distill_data}
\end{table}

\subsection{Experimental Setup}
\subsubsection{Training}
\label{sec:adapt_setup}

\paragraph{SFT} is conducted using AdamW optimizer with a learning rate of $2\times10^{-5}$, a batch size of 32 and sequence packing up to a total length of 16,384 tokens, and it is run for two epochs. 

\paragraph{OPD} uses AdamW optimizer with a learning rate of $1\times10^{-6}$. We set $\lambda = 0.25$, meaning that 25\% of training steps use student-generated (on-policy) sequences for distillation, while the remaining 75\% use sequences sampled from the reference dataset. OPD is performed for a single epoch. The OPD training is driven solely by the distillation loss with the forward KL divergence (teacher-to-student) as the distillation objective.

We use \texttt{Qwen/Qwen3-4B-Base} as the primary base model for our experiments. For RQ4, we use \texttt{ThaiLLM/ThaiLLM-8B} as the base model. All experiments are conducted on H100 GPUs. The 4B-scale experiments run on 4$\times$H100 GPUs and complete in under two days, while the 8B-scale experiments run on 8$\times$H100 GPUs and take approximately two days.

Full hyperparameters are provided in \Cref{tab:hyperparameters_sft_opd}.

\subsubsection{Evaluation}
\label{adapt:evaluation}

\paragraph{Benchmarks}

We evaluate all models on a diverse suite of benchmarks covering \textbf{chat}, \textbf{instruction following}, \textbf{knowledge}, \textbf{mathematical reasoning}, \textbf{code reasoning}, \textbf{tool use}, and \textbf{agentic retrieval}. We report results separately for English (EN), Thai (TH), and Thai code-switching (CS) when applicable.

\subparagraph{Chat}
We use \textbf{MT-Bench} as an LLM-as-a-judge evaluation of multi-turn conversational quality, focusing on helpfulness, correctness, and instruction adherence.
For English, we follow the standard LMSYS MT-Bench setup  \citep{zheng2023judgingllmasajudgemtbenchchatbot}.  
For Thai, we use \textbf{Thai MT-Bench} released by ThaiLLMLeaderboard  \citep{thaillm-leaderboard, payoungkhamdee2024mtbenchthai}, which is newly created in Thai and does not rely on translation; instead, the questions are written natively in a Thai linguistic and cultural context.  
Scores are reported separately for English (MTB EN) and Thai (MTB TH).

\subparagraph{Instruction following}
We evaluate verifiable instruction adherence using \textbf{IFEval}  \citep{zhou2023instructionfollowingevaluationlargelanguage}, reporting accuracy over predefined constraints.
We use both the original English IFEval and a Thai-translated version, \texttt{scb10x/ifeval-th}\footnote{\url{https://huggingface.co/datasets/scb10x/ifeval-th}}.
For the Thai setting, translations are manually verified to preserve instruction semantics, and the constraints are adapted natively to Thai.
Instruction-following results are reported separately for English (IFE EN) and Thai (IFE TH).

\subparagraph{Code-switching sampling robustness}
To measure robustness under language mixing in generation settings, we include a dedicated \textbf{Thai code-switching (CS)} evaluation.
We follow the evaluation protocol of \citet{pipatanakul2025adaptinglanguagespecificllmsreasoning}, which probes a model's ability to correctly follow instructions and generate coherent outputs using the characters commonly used by Thai speakers--namely Thai script with subtle English mixing.
The evaluation focuses on realistic Thai--English code-switching patterns that are prevalent in everyday communication, rather than artificial or rarely occurring language mixtures.
This metric is particularly sensitive to distributional mismatches in sampling and is therefore useful for diagnosing model brittleness.

\subparagraph{Knowledge benchmarks}
We evaluate factual and scientific knowledge using three benchmarks: (1) \textbf{GPQA Diamond}  \citep{rein2023gpqagraduatelevelgoogleproofqa}, a challenging multiple-choice benchmark in biology, physics, and chemistry, used in English only. (2) \textbf{MMLU Pro X (Thai)}  \citep{xuan2025mmluproxmultilingualbenchmarkadvanced}, a Thai-translated version of MMLU Pro  \citep{wang2024mmluprorobustchallengingmultitask}, evaluating broad academic knowledge. (3) \textbf{OpenThaiEval (OTE)}  \citep{yuenyong2025openthaigpt15thaicentricopen}, a native Thai benchmark covering Thai exam-style questions, natural language inference, and regional knowledge. These benchmarks are used to assess whether alignment methods preserve or degrade knowledge inherited from the base model.

\subparagraph{Mathematical reasoning}
We evaluate mathematical reasoning using \textbf{MATH500}  \citep{lightman2023letsverifystepstep}.
The benchmark consists of 500 English problems; we additionally include a Thai-translated version, resulting in 1{,}000 total problems.
Scores are reported separately for English (M500 EN) and Thai (M500 TH).

\subparagraph{Code reasoning}
We assess code generation and reasoning ability using \textbf{LiveCodeBench}  \citep{jain2024livecodebenchholisticcontaminationfree}, a contamination-resistant benchmark drawn from LeetCode, AtCoder, and Codeforces.
We use the \texttt{release\_v3} split containing 612 problems and report pass@1 accuracy (LCB EN).

\subparagraph{Tool use and agentic reasoning}
We evaluate tool use and multi-step agent with two benchmarks: (1) \textbf{BFCL v4}  \citep{patil2025bfcl}, a function-calling benchmark. We evaluate only the single-tool and multi-tool scenarios, comprising 4{,}441 problems in total. Results are reported as BFCL EN. (2) \textbf{HotpotQA (HPQA)}  \citep{yang2018hotpotqadatasetdiverseexplainable}, used as an end-to-end agentic retrieval benchmark. We evaluate on the medium subset, subsampled to 100 questions, using a simple ReAct-style agent connected to the Wikipedia API, following \citet{liu2024agentlite}. We report results separately for English (HPQA EN) and Thai (HPQA TH).

\paragraph{Metrics}
We use \textbf{accuracy} as our primary metric, where correctness is defined according to each dataset and evaluated using greedy decoding. The average score is computed as the unweighted mean across all reported benchmarks, with \texttt{MT-Bench} multiplied by \textit{10} to ensure equal weighting. Unless otherwise stated, higher scores indicate better performance.

\subsection{Results \& Discussion}
\label{sec:base2inst_eval}

To evaluate the effectiveness of our proposed recipe, we investigate whether post-trained models can achieve competitive performance with globally developed open-source models while remaining strong on local-language tasks under academic-scale resource constraints.
To this end, we assess the effectiveness of our recipe by ablating our design choices in RQ1–RQ3 and evaluate the generalization of the proposed recipe in RQ4.

\begin{enumerate}[label=\textbf{RQ\arabic*},leftmargin=2.5em,labelwidth=2.5em,labelsep=0.5em,itemsep=0pt,topsep=2pt]
    \item Does SFT alone, without OPD, suffice to achieve strong and robust performance?
    \item Is full-logits distillation truly necessary? Is it more effective than top-$K$ distillation across scenarios?
    \item Is a target-language dataset necessary at all post-training stages of our recipe to achieve strong performance on downstream tasks?
    \item Is our proposed recipe effective with a sovereignty-adapted base model?
\end{enumerate}

\subsubsection{A1: SFT Alone Is Insufficient to Achieve Strong and Robust Performance}
To answer the question of whether SFT alone is sufficient to achieve strong and robust performance without OPD, we use Qwen3 8B Instruct as our baseline model and compare a model fine-tuned using SFT alone with a model fine-tuned using our full recipe (SFT + OPD) across benchmarks. \Cref{tab:sft_q1} suggests that \emph{SFT alone is \textbf{not sufficient}} to achieve strong overall performance across our evaluation suite, as it yields a lower average performance compared to the SFT+OPD model. While SFT produces a functional instruction-following model, it substantially underperforms on several capability axes and exhibits clear brittleness under distribution shift. The degradation is most visible in code-switching and tool use: the SFT model drops sharply on Thai code-switching (CS: 65.4 vs. 93.4 from the SFT+OPD model) and achieves zero scores on HPQA in both languages. This behavior is consistent with the fact that standard standard SFT maximizes the likelihood of the observed target token but does not explicitly encourage probability mass over other plausible or semantically equivalent tokens, resulting in a sharply peaked and brittle token distribution \citep{kim-rush-2016-sequence}.

In contrast, our recipe (SFT+OPD) improves performance over SFT across most benchmarks, raising the average score by +6.49 points (37.45 $\rightarrow$ 43.94). We hypothesize that these gains come from two factors. First, distillation provides a richer training signal than cross-entropy against a single ground-truth sequence by leveraging the full distribution of the teacher \citep{hinton2015distillingknowledgeneuralnetwork}. Second, because distillation is performed on-policy, the training targets are computed on student-generated trajectories, reducing mismatch between training contexts and the model's current behavior \citep{agarwal2024onpolicydistillationlanguagemodels}. Together, these factors are potentially improve generalization on open-ended generation tasks, consistent with gains on MT-Bench, MATH500, HotpotQA, and code-switching. In contrast, performance on knowledge-focused benchmarks (e.g., MMLU, OpenThaiEval, and GPQA) remains similar to the SFT-only model, suggesting these capabilities are largely inherited from the base model and that OPD is less effective at improving them.

\begin{table}[!ht]
\tabcolsep=1mm
\centering
\renewcommand*{\arraystretch}{1.2}
\resizebox{\linewidth}{!}{
\begin{tabular}{l|cc|c|cc|ccc|cc|c|ccc|c}
\toprule
    \multirow{3}{*}{\textbf{Model}} &
    \multicolumn{2}{c|}{\textbf{Chat}} &
    \multicolumn{1}{c|}{\textbf{CS}} &
    \multicolumn{2}{c|}{\textbf{IF}} &
    \multicolumn{3}{c|}{\textbf{Knowledge}} &
    \multicolumn{2}{c|}{\textbf{Math}} &
    \multicolumn{1}{c|}{\textbf{Code}} &
    \multicolumn{3}{c|}{\textbf{Tool / Agent}} &
    \multicolumn{1}{c}{\multirow{3}{*}{\textbf{Average}}} \\
    & EN & TH & TH & EN & TH & EN & \multicolumn{2}{c|}{TH} & EN & TH & EN & EN & TH & EN &  \\
    & MTB & MTB & CS & IFE & IFE & GPQA & MMLU & OTE & M500 & M500 & LCB & HPQA & HPQA & BFCL & \\
\midrule
    Qwen3 Instruct & 8.65 & 7.20 & 96.20 & 87.42 & 79.68 & 39.39 & 34.18 & 67.23 & 87.20 & 78.76 & 33.50 & 14.00 & 8.00 & 31.53 & 48.07 \\
    \midrule
    SFT & 7.20 & 5.67 & 65.40 & 70.94 & 73.35 & 29.29 & \textbf{31.97} & \textbf{61.28} & 50.40 & \textbf{70.94} & \textbf{33.50} & 0.00 & 0.00 & 24.36 & 37.45 \\
    SFT+OPD & \textbf{8.24} & \textbf{6.44} & \textbf{93.40} & \textbf{83.18} & \textbf{75.98} & \textbf{33.84} & 29.76 & 61.19 & \textbf{74.00} & 66.93 & 31.21 & \textbf{15.00} & \textbf{9.00} & \textbf{26.95} & \textbf{43.94} \\
\bottomrule
\end{tabular}
}
\caption{Performance comparison of the instruction-tuned baseline model, SFT, and SFT with on-policy distillation (SFT+OPD) across chat, instruction following, knowledge, math reasoning, code, and tool-use benchmarks in English and Thai. Results show that SFT alone underperforms the base model and suffers from poor generalization, while on-policy distillation consistently improves robustness and overall performance.}
\label{tab:sft_q1}
\end{table}

\subsubsection{A2: Full-Logits Distillation Is Not Always Necessary, but It Improves Sampling Robustness}\label{sec:full_vs_topk_opd}
We compare the SFT+OPD (OPD Full) model from the previous experiment, which uses full-logits distillation, with a model trained using top-$K$ OPD (OPD Top-$K$) initialized from the SFT-only model in the previous section, in order to examine the differences between the two OPD paradigms.
As shown in \Cref{tab:sft_q2}, full-logits distillation achieves a higher average score (43.94 vs.\ 42.81), although the differences are not consistent across tasks. The most pronounced gap appears in Thai code-switching, where OPD with full logits substantially outperforms OPD Top-$K$ (CS: 93.4 vs.\ 69.8). In contrast, several benchmarks show comparable performance or even slight improvements under Top-$K$ distillation, particularly on tasks with a single correct answer, MATH500 and HotpotQA.

These results suggest that \textbf{full-logits distillation} provides better \emph{robustness under sampling in open-ended generation}, especially for multilingual and less well-represented languages. We hypothesize that the full teacher distribution provides richer token-level probabilities over \emph{long-tail tokens}, (e.g., linguistically plausible alternatives, transliterations, and mixed-language function words), which helps reduce spurious character-level errors during stochastic decoding and is often lost with Top-$K$ distillation.

However, Top-$K$ distillation appears sufficient for tasks with more constrained output spaces and single-correct-answer objectives, where the primary challenge lies in evidence selection and reasoning rather than modeling fine-grained token-level uncertainty. In summary, while full-logits distillation is not strictly necessary to achieve reasonable overall performance, it provides a clear advantage for long-tail tokens robustness. We therefore adopt full-logits distillation for the rest of the experiments, as it enables higher accuracy on code-switching tasks and is more appropriate for a sovereignty-focused setting.

\begin{table}[!ht]
\tabcolsep=1mm
\centering
\renewcommand*{\arraystretch}{1.2}
\resizebox{\linewidth}{!}{
\begin{tabular}{l|cc|c|cc|ccc|cc|c|ccc|c}
\toprule
    \multirow{3}{*}{\textbf{Model}} &
    \multicolumn{2}{c|}{\textbf{Chat}} &
    \multicolumn{1}{c|}{\textbf{CS}} &
    \multicolumn{2}{c|}{\textbf{IF}} &
    \multicolumn{3}{c|}{\textbf{Knowledge}} &
    \multicolumn{2}{c|}{\textbf{Math}} &
    \multicolumn{1}{c|}{\textbf{Code}} &
    \multicolumn{3}{c|}{\textbf{Tool / Agent}} &
    \multicolumn{1}{c}{\multirow{3}{*}{\textbf{Average}}} \\
    & EN & TH & TH & EN & TH & EN & \multicolumn{2}{c|}{TH} & EN & TH & EN & EN & TH & EN &  \\
    & MTB & MTB & CS & IFE & IFE & GPQA & MMLU & OTE & M500 & M500 & LCB & HPQA & HPQA & BFCL & \\
    \midrule
    OPD Full & \textbf{8.24} & \textbf{6.44} & \textbf{93.40} & 83.18 & 75.98 & \textbf{33.84} & \textbf{29.76} & 61.19 & 74.00 & 66.93 & \textbf{31.21} & 15.00 & 9.00 & \textbf{26.95} & \textbf{43.94} \\
    OPD Top-$K$ & \textbf{8.24} & 6.38 & 69.80 & \textbf{83.55} & \textbf{76.70} & 31.31 & 28.91 & \textbf{62.13} & \textbf{74.60} & \textbf{68.34} & 30.07 & \textbf{20.00} & \textbf{14.00} & 25.36 & 42.81 \\
\bottomrule
\end{tabular}
}
\caption{Comparison between OPD using full-logits distillation and Top-$K$ distillation across the evaluation suite. Full-logits distillation achieves a higher overall average and substantially improves robustness on Thai code-switching, while Top-$K$ distillation performs comparably on more constrained, single-correct-answer tasks such as math reasoning and HotpotQA.}
\label{tab:sft_q2}
\end{table}

\subsubsection{A3: Target-Language Data Is Essential for SFT, While Primarily Benefits OPD In Thai Native Tasks, With Limited Impact on Thai Translated Tasks}
To assess the impact of the target-language dataset at each stage of the training recipe, we compare models trained at each stage with and without the target-language data, with Thai consistently excluded across the SFT and OPD training stages in the latter setting.
\Cref{tab:sft_q3} shows that the inclusion of Thai target-language dataset plays different roles for each training stage.

For \textbf{SFT}, Thai data is clearly \emph{important}. Removing Thai data (SFT w/o Thai) causes large regressions across Thai-facing evaluations, including Thai chat (MTB TH: 5.67 $\rightarrow$ 4.36), Thai instruction following (IFE TH: 73.35 $\rightarrow$ 57.44), and especially code-switching (CS: 65.4 $\rightarrow$ 34.4). These results indicate that SFT is highly sensitive to data coverage: without explicit Thai supervision, the model fails to learn proper Thai alignment. Interestingly, adding Thai data also benefits some English benchmarks, suggesting that broader multilingual instruction coverage may improve overall alignment and robustness.

For \textbf{OPD}, the gap between using and removing Thai data is \emph{smaller} in terms of absolute average score (43.94 vs.\ 42.02), but the effect is more targeted. Thai data primarily improves \emph{Thai Native} tasks--Thai chat (MTB TH: 6.28 $\rightarrow$ 6.44), OpenThaiEval (59.32 $\rightarrow$ 61.19), and code-switching (80.8 $\rightarrow$ 93.4), as well as instruction-following (IFE TH: 72.62 $\rightarrow$ 75.98), which we attribute to greater diversity in constraint scenarios. In contrast, Thai translated or language-agnostic tasks such as MATH500 and MMLU Pro X show little change. English-only evaluations and knowledge-heavy multiple-choice benchmarks remain largely stable (e.g., GPQA).

Overall, this pattern suggests that OPD already transfers much of the general interaction behavior from the teacher even without Thai data, while Thai supervision remains valuable for refining Thai-specific language capabilities.

\begin{table}[!ht]
\tabcolsep=1mm
\centering
\renewcommand*{\arraystretch}{1.2}
\resizebox{\linewidth}{!}{
\begin{tabular}{l|cc|c|cc|ccc|cc|c|ccc|c}
\toprule
    \multirow{3}{*}{\textbf{Model}} &
    \multicolumn{2}{c|}{\textbf{Chat}} &
    \multicolumn{1}{c|}{\textbf{CS}} &
    \multicolumn{2}{c|}{\textbf{IF}} &
    \multicolumn{3}{c|}{\textbf{Knowledge}} &
    \multicolumn{2}{c|}{\textbf{Math}} &
    \multicolumn{1}{c|}{\textbf{Code}} &
    \multicolumn{3}{c|}{\textbf{Tool / Agent}} &
    \multicolumn{1}{c}{\multirow{3}{*}{\textbf{Average}}} \\
    & EN & TH & TH & EN & TH & EN & \multicolumn{2}{c|}{TH} & EN & TH & EN & EN & TH & EN &  \\
    & MTB & MTB & CS & IFE & IFE & GPQA & MMLU & OTE & M500 & M500 & LCB & HPQA & HPQA & BFCL & \\
    \midrule
    SFT & \textbf{7.20} & \textbf{5.67} & \textbf{65.40} & \text{70.94} & \textbf{73.35} & 29.29 & \textbf{31.97} & 61.28 & 50.40 & \textbf{70.94} & \textbf{33.50} & 0.00 & \textbf{0.00} & 24.36 & \textbf{37.45} \\
    SFT w/o Thai & 7.14 & 4.36 & 34.40 & 66.33 & 57.44 & \textbf{35.35} & 30.44 & \textbf{62.13} & \textbf{51.20} & 57.31 & 31.21 & \textbf{1.00} & \textbf{0.00} & \textbf{24.70} & 33.07 \\
    \midrule
    OPD & 8.24 & \textbf{6.44} & \textbf{93.40} & \textbf{83.18} & \textbf{75.98} & \textbf{33.84} & 29.76 & \textbf{61.19} & \textbf{74.00} & 66.93 & \textbf{31.21} & \textbf{15.00} & \textbf{9.00} & 26.95 & \textbf{43.94} \\
    OPD w/o Thai & \textbf{8.26} & 6.28 & 80.80 & 82.04 & 72.62 & 32.32 & \textbf{32.65} & 59.32 & 73.60 & \textbf{69.34} & 29.90 & 8.00 & 6.00 & \textbf{27.19} & 42.02 \\
\bottomrule
\end{tabular}
}
\caption{Effect of Thai general-data inclusion on SFT and OPD. Removing Thai data severely degrades SFT performance on Thai-facing and code-switching tasks, while OPD is more robust and shows smaller overall regressions, with Thai data primarily improving Thai-specific and Thai-adjacent evaluations.}
\label{tab:sft_q3}
\end{table}

\subsubsection{A4: Our Recipe Is Effective With a Sovereignty-Adapted Base Model}

In all preceding experiments (RQ1-3), our recipe--SFT followed by OPD--has shown strong effectiveness in adapting Qwen3 8B Instruct. However, an open question remains: how effective is our recipe when applied to a sovereignty-adapted base model?

To answer this question, we apply our recipe to \texttt{ThaiLLM-8B}\footnote{\url{https://huggingface.co/ThaiLLM/ThaiLLM-8B}}, a sovereign base model from the \texttt{ThaiLLM}\footnote{\url{https://huggingface.co/ThaiLLM}} project obtained by continuing pretraining \texttt{Qwen3-8B-Base} on 64B tokens of Thai corpus. Because \texttt{ThaiLLM-8B} is a base model, it \emph{does not} natively exhibit instruction-following behavior, making it a strong test of whether our method can reliably produce an instruction-tuned model while preserving local-language strengths. We release the resulting model as \texttt{\textbf{Typhoon-S-ThaiLLM-8B}}\footnote{\url{https://huggingface.co/scb10x/typhoon-s-thaillm-8b-instruct-research-preview}}\footnote{To further improve performance, we replace the teacher model with a stronger one \citep{hinton2015distillingknowledgeneuralnetwork}: \href{https://huggingface.co/Qwen/Qwen3-30B-A3B-Instruct-2507}{\texttt{Qwen/Qwen3-30B-A3B-Instruct-2507}}.}. Because \texttt{ThaiLLM-8B} is a \emph{base} model and does not reliably follow instructions, we do not report its scores on instruction-tuned benchmarks; instead, we evaluate the instruction-tuned model produced by our pipeline (\texttt{Typhoon-S-8B}) against a strong instruction-tuned baseline (\texttt{Qwen3-8B}).

\begin{table}[!h]
\tabcolsep=1mm
\centering
\small
\renewcommand*{\arraystretch}{1.2}
\begin{tabular}{l|c|c|c|c|c|c}
\toprule
    \multirow{2}{*}{\textbf{Model}} & \textbf{Chat} & \textbf{CS} & \textbf{IF} & \textbf{Knowledge} & \textbf{Agent} & \multirow{2}{*}{\textbf{Average}} \\
     & MTB & CS & IFE & OTE &  HPQA & \\
    \midrule
    Qwen3-8B & 7.08 & 95.40 & \textbf{80.47} & 63.66 & 23.00 & 66.66 \\
    \textbf{Typhoon-S-8B} & \textbf{7.89} & \textbf{96.60} & 76.45 & \textbf{67.06} & \textbf{37.00} & \textbf{71.20} \\
\bottomrule
\end{tabular}
\caption{
Thai-only evaluation (higher is better) comparing \texttt{Typhoon-S-8B}—trained from the sovereignty-focused base \texttt{ThaiLLM-8B} using our SFT+OPD (full-logits) recipe—against \texttt{Qwen3-8B}. \texttt{Typhoon-S-8B} improves Thai chat, code-switching, Thai knowledge (OTE), and agentic QA, yielding a higher Thai average (71.20 vs.\ 66.66).
}
\label{tab:sft_q51}
\end{table}

\Cref{tab:sft_q51} shows that the resulting instruction-tuned model (\texttt{Typhoon-S-8B}\footnote{For conciseness, we use \texttt{Typhoon-S-ThaiLLM-8B} interchangeably with \texttt{Typhoon-S-8B} throughout the rest of the paper.}) surpasses \texttt{Qwen3-8B} on the \emph{strict} Thai evaluation suite (i.e., tasks written originally in Thai rather than translated) and Agentic task. In particular, \texttt{Typhoon-S-8B} improves Thai chat quality, code-switching robustness, OpenThaiEval, and retrieval-style QA, yielding a higher overall Thai average (71.20 vs.\ 66.66). This suggests that starting from a Thai-optimized base model and applying our post-training recipe can produce a model that is both aligned and strong on local-language tasks. An interesting observation is that, on agentic tasks, performance improves to some extent when our method is applied to a sovereignty-focused model. This finding raises the question of whether \emph{local knowledge alone is sufficient} to improve performance in low-resource languages under unbiased post-training, which we leave for future work.

\begin{table}[!t]
\tabcolsep=1mm
\centering
\renewcommand*{\arraystretch}{1.2}
\resizebox{\linewidth}{!}{
\begin{tabular}{l|cc|c|cc|ccc|cc|c|ccc|c}
\toprule
    \multirow{3}{*}{\textbf{Model}} &
    \multicolumn{2}{c|}{\textbf{Chat}} &
    \multicolumn{1}{c|}{\textbf{CS}} &
    \multicolumn{2}{c|}{\textbf{IF}} &
    \multicolumn{3}{c|}{\textbf{Knowledge}} &
    \multicolumn{2}{c|}{\textbf{Math}} &
    \multicolumn{1}{c|}{\textbf{Code}} &
    \multicolumn{3}{c|}{\textbf{Tool / Agent}} &
    \multicolumn{1}{c}{\multirow{3}{*}{\textbf{Average}}} \\
    & EN & TH & TH & EN & TH & EN & \multicolumn{2}{c|}{TH} & EN & TH & EN & EN & TH & EN &  \\
    & MTB & MTB & CS & IFE & IFE & GPQA & MMLU & OTE & M500 & M500 & LCB & HPQA & HPQA & BFCL & \\
    \midrule
    Qwen3-8B & \textbf{8.69} & 7.08 & 95.40 & \textbf{87.64} & \textbf{80.47} & \textbf{41.41} & \textbf{42.18} & 63.66 & \textbf{81.20} & \textbf{73.95} & \textbf{63.39} & \textbf{52.00} & 23.00 & \textbf{36.17} & \textbf{54.02} \\
    \textbf{Typhoon-S-8B} & 8.19 & \textbf{7.89} & \textbf{96.60} & 79.28 & 76.45 & 31.31 & 34.18 & \textbf{67.06} & 71.60 & 65.93 & 33.49 & 47.00 & \textbf{37.00} & 27.39 & 49.99 \\
\bottomrule
\end{tabular}
}
\caption{Full-suite evaluation in English and Thai comparing \texttt{Typhoon-S-8B} (SFT+OPD with full-logits on the sovereignty-focused base \texttt{ThaiLLM-8B}) against \texttt{Qwen3-8B}. \texttt{Typhoon-S-8B} is stronger on Thai-centric benchmarks (e.g., MT-Bench TH, OpenThaiEval, and HPQA TH), but lags on English scientific/knowledge benchmarks (GPQA, MMLU) and mathematics (MATH500), resulting in a slightly lower overall average (51.88 vs.\ 54.72). Scores are reported per benchmark; higher is better unless otherwise noted.}
\label{tab:sft_q52}
\end{table}

When evaluated on the full English+Thai benchmark suite (Table~\ref{tab:sft_q52}), \texttt{Typhoon-S-8B} reaches performance comparable to \texttt{Qwen3-8B} overall, though a gap remains on two categories: (1) hard/scientific knowledge (e.g., GPQA and MMLU) and (2) mathematics (MATH500) (3) coding (LiveCodeBench). We also observe that several translated benchmarks may be biased toward Qwen3 8B, which is a highly multilingual model (119 languages), likely due to cross-lingual ability with English-centric data. In contrast, on sovereignty-relevant evaluations emphasizing native Thai usage and local knowledge (e.g., OpenThaiEval and MT-Bench TH), the sovereignty-focused model maintains a clear advantage over the highly multilingual model. Overall, these results show that our method adapts well to a sovereignty-focused base model, yielding a strong instruction-following model while preserving local-language performance.

\section{Sovereign Capability}
\label{sec:frontier}

Although the recipe introduced in the previous section is effective at transforming general-purpose base models into instruction-tuned models with strong performance across a wide range of Thai and English tasks, such models often fall short on more complex problems—especially those requiring regional, cultural, or domain-specific knowledge and reasoning that are underrepresented in general-domain data. A common approach to mitigating this limitation is to introduce additional post-training stages targeted at specific capabilities. Reinforcement-learning-based approaches, such as RFT \citep{deepseekr1_2025}, have recently become a standard choice for this purpose, as they enable targeted optimization for specific tasks while offering better generalization than many alternative post-training techniques \citep{chu2025sft,zhu2025the,chen2025retainingdoingroleonpolicy,shenfeld2025rls}.

However, standard RFT pipelines are typically designed to optimize task performance under fixed data and interaction assumptions \citep{deepseekr1_2025, zhao2025echochamberrlamplify}. As a result, they exhibit two important limitations in domain-specialized settings: they are often \emph{ineffective at introducing new domain knowledge} that is absent from the base model, and they provide \emph{limited support for agentic reasoning behaviors} such as multi-turn decision-making and tool use. These limitations motivate the need for extensions to the standard RFT recipe that explicitly target both knowledge injection and agentic capabilities.

In this section, we introduce two such extensions to a standard RFT pipeline, both built on GRPO \citep{shao2024deepseekmathpushinglimitsmathematical}. Specifically, we propose \textbf{Injected Knowledge GRPO (InK-GRPO)}, which augments the GRPO objective with stochastic next-token prediction, and \textbf{Agentic RFT}, which enables multi-step reasoning with external tools in a controlled retrieval-augmented environment. 
We describe each extension in turn.

\subsection{Injected Knowledge GRPO (InK-GRPO)}
\label{sec:rl_pretrain}

Prior work suggests that RFT primarily amplifies knowledge already present in a model, rather than introducing new factual information \citep{zhao2025echochamberrlamplify}. This limitation is especially problematic for sovereign and domain-specialized settings, where relevant domain knowledge and region-specific content are often underrepresented in both the base model and general-domain instruction data.

To mitigate this issue, we propose \textbf{InK-GRPO}, which augments the standard GRPO objective with an auxiliary \emph{next-token prediction} loss computed on a separate in-domain text corpus. The key idea is to stochastically inject domain knowledge during RFT while preserving the benefits of task-focused reinforcement learning.

\paragraph{Data Sources} InK-GRPO training uses two distinct data sources:

\begin{enumerate}
    \item \textbf{Task prompt datasets}: Used to generate GRPO rollouts and compute the GRPO loss ($\mathcal{L}_{\text{GRPO}}$)
    \item \textbf{In-domain text corpus}: Used to compute the auxiliary next-token prediction loss ($\mathcal{L}_{\text{CE}}$)
\end{enumerate}

\paragraph{Training Procedure} At each optimization step, we stochastically apply an auxiliary cross-entropy (CE) loss with probability $\rho$. Formally, the training objective is:

\begin{equation}
\label{eq:training_loss}
    \mathcal{L} = \mathcal{L}_{\text{GRPO}} + \lambda \, b \, \mathcal{L}_{\text{CE}}\text{,}
\end{equation}

\noindent where $b \sim \text{Bernoulli}(\rho)$ determines whether the auxiliary loss is applied, and $\lambda$ controls the relative weight of $\mathcal{L}_{\text{CE}}$ when active. This stochastic scheduling allows the model to alternate between domain knowledge acquisition (via $\mathcal{L}_{\text{CE}}$) and task-specific optimization (via $\mathcal{L}_{\text{GRPO}}$), while maintaining focus on the latter through standard RFT. Pseudo-code for InK-GRPO is provided in \Cref{appendix:rl_pretrain_code}.

\subsection{Agentic RFT}
\label{sec:agentic_rft}

While InK-GRPO addresses limitations in domain knowledge acquisition, standard RFT pipelines remain limited in their ability to support agentic reasoning behaviors such as multi-turn decision-making and tool use. To address this complementary limitation, we explore training an agentic reasoning model using RFT, with final-answer accuracy as the primary reward signal.

During both training and inference, the model interacts with a controlled retrieval-augmented generation (RAG) \emph{environment} through two tools: (1) \texttt{search}, which performs semantic retrieval over an in-domain corpus, (2) \texttt{read}, which returns the full in-domain document given its identifier. The model learns to interleave natural language reasoning with tool calls across multiple interaction turns in order to gather relevant information and produce a final answer.

In this setup, GRPO is applied over entire interaction trajectories, enabling joint optimization of reasoning steps, tool-use decisions, and final-answer correctness within a unified RFT framework. Since tool outputs are not generated by the model, tool responses are masked during gradient computation.

\paragraph{Environment}

We formulate the RAG setup as a \emph{partially observable} environment in which the agent incrementally acquires information through tool interactions. At each step, the agent produces a reasoning trace and then either issues a tool call or emits a final answer. Observations returned by the environment consist of retrieved document identifiers in response to \texttt{search} actions, or full document contents in response to \texttt{read} actions. An episode terminates when the agent outputs a final answer or performs no further tool calls, at which point the reward is computed based on answer accuracy.

\subsection{Experimental Setup}
\label{sec:frontier_setup}

\subsubsection{Training}

We follow a standard RFT setup similar to DeepSeek-R1 \citep{deepseekr1_2025}. Specifically, we use GRPO as the base optimization algorithm integrated with our two aforementioned extensions: \texttt{InK-GRPO} and \texttt{Agentic Training}. Our training framework is built on veRL \citep{sheng2024hybridflow}.

\paragraph{Reward Function}
Our reward function consists of two components: a format reward and an accuracy reward.
\begin{itemize}
    \item \textbf{Format:} We apply a format reward to encourage the model to place its reasoning within the \texttt{<thinking></thinking>} tags, with the final response generated outside these tags. We use the \texttt{<thinking>} tag instead of \texttt{<think>} because \texttt{<think>} is a special token in the Qwen3 model but is not learned by the Instruct model. This results in uninitialized embeddings, which substantially increases training resource requirements.
    \item \textbf{Accuracy:} We evaluate the correctness of the generated response by comparing it to a reference answer. Due to the open-ended nature of the outputs, we employ an LLM-as-a-Judge approach \citep{taveekitworachai2026robustnessanswerformatsmedical} to determine answer correctness. We use a reasoning-based judge model (e.g., \texttt{gpt-5-nano}) for correctness evaluation. In pilot experiments, instruction-tuned judge models (e.g., \texttt{gpt-4o-mini}) led to reward hacking, incentivizing the agent to generate excessively long responses; this behavior is mitigated by switching to a reasoning judge model.
\end{itemize}

The final reward ($r \in [0, 1]$) is a weighted sum of the accuracy ($r_{\text{acc}}$) and format ($r_{\text{format}}$) rewards. The raw accuracy score is assigned in the range $[0, 2]$, where 0 indicates an incorrect answer, 1 a partially correct answer, and 2 a fully correct answer. This score is then normalized to $[0, 1]$, while the format reward is binary in $[0, 1]$. The weights of accuracy and format rewards are 0.9 and 0.1, respectively. This weighting follows our prior work in RFT \citep{taveekitworachai2025priorpromptengineeringreinforcement, taveekitworachai2026robustnessanswerformatsmedical}. Formally, the reward function is

\begin{equation}
    r = 0.9 \times \frac{r_{\text{acc}}}{2} + 0.1 \times r_{\text{format}}
\end{equation} 

For agentic RFT, we use only the accuracy reward, i.e., we do not apply any format reward, as text-before tool calling already serves as implicit reasoning. We do not incorporate rewards for successful tool calls or any other complex reward components.

\paragraph{Training Data}
The RAG environment operates over a static in-domain document collection indexed as a vector database. Documents are embedded using \texttt{Qwen/Qwen3-Embedding-0.6B}\footnote{\url{https://huggingface.co/Qwen/Qwen3-Embedding-0.6B}} and stored in a FAISS index with an IVF-SQ8 configuration. For each \texttt{search} action, the environment returns the top three documents based on vector similarity, while the \texttt{read} action returns the full content of a specified document. No additional re-ranking or post-processing is applied in the retrieval pipeline.

As InK-GRPO requires two sources of training data, \textbf{RFT data} and \textbf{CE data}, we prepare them as follows:

\begin{itemize}
    \item \textbf{RFT}: Each sample is structured as a question-answering task, where the model is prompted with a question, generates responses, and receives reward feedback by comparing its generated response against a reference answer
    \item \textbf{CE}: Each sample consists of unstructured text obtained by each in-domain textual evidence from the dataset, without including the question itself. This form of semi-supervision allows the model to acquire new knowledge \citep{howllmacquirefactualknowledge} without being forced to generate a specific response, as in supervised learning. At the same time, the model already possesses the knowledge required to answer the question, which RFT should focus on identifying and effectively utilizing.
\end{itemize}

We primarily conduct our training and evaluation on \textbf{NitiBench} \citep{akarajaradwong-etal-2025-nitibench}, a Thai legal dataset that represents a sovereignty-focused setting. As NitiBench is provided as a document collection suitable for RAG use cases (i.e., context-query-answer triples), we construct the training dataset by leveraging the contextual information associated with each example while omitting this context from the query itself.

For NitiBench, we conduct experiments on the \texttt{ccl} split and treat the \texttt{positive\_contexts} field as in-domain text for the CE loss computation, while the \texttt{question} and \texttt{positive\_answer} fields are used for the GRPO loss computation. To evaluate the generalizability of our approach beyond a single domain, we additionally conduct experiments in RQ5 on \textbf{MIRAGE-Bench} \citep{thakur2025miragebenchautomaticmultilingualbenchmark}, a multilingual RAG benchmark designed to cover broader and more general settings. In this case, we use the training split of \texttt{MIRAGE-Bench-Instruct}\footnote{\url{https://huggingface.co/datasets/nthakur/mirage-bench-instruct}} and focus on the Thai subset of the benchmark.

To avoid potential interference between domains, each model is trained separately on a single dataset, ensuring that injected in-domain knowledge remains consistent and non-conflicting.

\paragraph{Training Details}

For our main experiments, we use Qwen3-4B-Instruct-2507  \citep{yang2025qwen3technicalreport} as the base model. We perform full fine-tuning using AdamW optimizer with a learning rate of $1\times10^{-6}$. The sampling temperature is 0.7. We using \texttt{gpt-5-nano} as the judge model for all experiments. The full hyperparameters is in \Cref{tab:hyperparameters_rl}.

\subparagraph{GRPO Loss} We adopt the DAPO's decoupled-clip-higher strategy \citep{yu2025dapoopensourcellmreinforcement}, setting the high clip ratio to $0.24$ and the low clip ratio to $0.20$. This choice is motivated by our observation of entropy collapse when using the standard single clipping value (0.20) in pilot experiments. We also use overlong reward shaping \citep{yu2025dapoopensourcellmreinforcement} to penalize excessively long responses and getting cut off.

\subparagraph{CE Loss} We set $\rho$ to 0.6 and $\lambda$ to 0.1, which were selected empirically based on pilot experiments. We did not perform a dedicated ablation study on these hyperparameters; a systematic investigation of their impact is left for future work. The batch size for calculating CE loss is set to twice the mini-batch size, and the updates are performed at the mini-batch level.

\subsubsection{Evaluation}

\paragraph{Benchmarks}
We evaluate our methods on benchmarks that mirror the training environments and emphasize underrepresented languages and domain-specific knowledge, settings in which base models typically underperform. Our primary evaluation benchmark is \textbf{NitiBench} \citep{akarajaradwong-etal-2025-nitibench}, which targets Thai legal reasoning and requires both precise reasoning and knowledge of local law. This benchmark serves as the main testbed for assessing the benefits of InK-GRPO and its integration with agentic RFT in sovereign settings.

To assess generalization beyond a single domain, we additionally conduct RQ5 experiments on \textbf{MIRAGE-Bench} \citep{thakur2025miragebenchautomaticmultilingualbenchmark}, a Wikipedia-based multilingual RAG benchmark derived from TyDi QA and MIRACL \citep{clark-etal-2020-tydi,zhang-etal-2023-miracl}. These evaluation settings align with the corresponding training datasets and allow us to isolate generalization effects without introducing mismatched retrieval domains.

To reduce inference costs, we randomly sample 10\% of the evaluation data (approximately $n=300$), which pilot experiments show yields comparable results. For all evaluations, we use greedy decoding.

\paragraph{Metrics}
We evaluate each test split corresponding to the training datasets and report task-specific \textbf{accuracy}. In RQ8, general performance is measured using the metrics defined in \Cref{adapt:evaluation}. To quantify general capability retention after domain-specific fine-tuning, we report general performance as the unweighted average across all evaluations described in \Cref{adapt:evaluation}, ensuring equal weight across tasks.

\subsection{Results \& Discussion}
\label{sec:frontier_eval}

To evaluate the effectiveness of our method, we seek to answer the following research questions:

\begin{enumerate}[start=5,label=\textbf{RQ\arabic*},leftmargin=2.5em,labelwidth=2.5em,labelsep=0.5em,itemsep=0pt,topsep=2pt]
    \item Does InK-GRPO improve performance over GRPO in sovereignty-focused settings?
    \item In InK-GRPO, is pretraining-style data or SFT-style instruction data more effective for CE loss?
    \item Does agentic RFT further improve performance, and can the InK-GRPO extend to modern multi-turn agentic RFT settings?
    \item How does InK-GRPO affect general capabilities and catastrophic forgetting?
\end{enumerate}

\subsubsection{A5: InK-GRPO Improves Performance in Sovereignty-Focused Settings}

We examine whether InK-GRPO--by jointly optimizing a task performance objective and a pre-training (next-token prediction) objective during RFT--enhances sovereignty-focused, task-specific performance. We compare InK-GRPO with standard GRPO on both NitiBench and MIRAGE-Bench. As described in our training setup, we train separate models for each dataset, resulting in four models in total.

\Cref{tab:rl_q1} shows that InK-GRPO variants outperform their GRPO-only counterparts on both datasets. On NitiBench, InK-GRPO yields a 4\% absolute improvement in accuracy (19.30\% vs.\ 15.82\%), while on MIRAGE-Bench, it achieves a consistent gain of 1.6\% (22.63\% vs.\ 20.99\%). These results indicate that augmenting RFT with a stochastic in-domain next-token prediction objective, as in InK-GRPO, yields consistent performance gains.

We further observe that the magnitude of improvement varies across datasets due to their differing characteristics. \textbf{NitiBench}, which emphasizes legal reasoning and multi-step reasoning, benefits substantially from GRPO itself--showing more than a 100\% relative improvement over the base model. In contrast, \textbf{MIRAGE-Bench}, which focuses on multilingual understanding, exhibits smaller but consistent gains. Nonetheless, the additional improvement achieved by InK-GRPO indicates that integrating in-domain knowledge positively influences both datasets, regardless of task type.

Finally, We further compare against GPT-5, which remains substantially stronger across both benchmarks, largely due to its much larger model scale compared to our 4B backbone. The gap is especially pronounced on \textbf{MIRAGE-Bench}, which emphasizes general and multilingual knowledge, while on \textbf{NitiBench} GPT-5 exhibits a relative performance drop, reflecting the increased difficulty of sovereignty-focused reasoning tasks. These results highlight the need for targeted adaptation methods such as InK-GRPO for improving performance in specialized domains.

\begin{table}[ht]
\centering
\renewcommand*{\arraystretch}{1.2}
\small
\begin{tabular}{l|cc}
\hline
\textbf{Model} & \textbf{NitiBench Accuracy} & \textbf{MIRAGE Accuracy}  \\
\hline
GPT-5 & \textbf{28.79\%} & \textbf{53.91\%} \\
\midrule
Qwen3 4B Instruct 2507 & 5.90\% & 17.70\% \\
\midrule
GRPO & 15.82\% & 20.99\% \\
\textbf{InK-GRPO} & 19.30\% & 22.63\% \\
\hline
\end{tabular}
\caption{
Comparison of GRPO and InK-GRPO trained separately on NitiBench and MIRAGE-Bench. InK-GRPO consistently improves performance across datasets.
}
\label{tab:rl_q1}
\end{table}

\subsubsection{A6: Pretraining In-Domain Data Is More Effective Than SFT Data For InK-GRPO CE Loss}

Since the InK-GRPO CE loss is computed from a training source separate from that used for the GRPO objective, we investigate whether the CE data should be drawn from pretraining-style data (as in RQ6) or SFT-style instruction data. To this end, we construct an SFT-style variant by using the \texttt{question} field as the prompt and the \texttt{positive\_answer} field as the target the NitiBench training set. As shown in \Cref{tab:rl_q4}, both InK-GRPO variants outperform the GRPO baseline on NitiBench. However, InK-GRPO trained with \textbf{pretraining-style data} achieves the highest accuracy (19.30\%), whereas the variant trained with SFT-style data yields a smaller improvement (16.89\%).

These results suggest that, for joint optimization during RFT, \textbf{pretraining-style in-domain text is more effective} than SFT-style instruction data. One possible explanation is that SFT data may overly constrain the policy by biasing it toward a narrow set of preferred responses, thereby limiting the exploration necessary for effective RFT. In contrast, pretraining-style language modeling offers broader domain exposure without directly optimizing the same behavioral space, which may better preserve exploration while still injecting domain knowledge.

\begin{table}[ht]
\centering
\renewcommand*{\arraystretch}{1.2}
\small
\begin{tabular}{l|c}
\hline
\textbf{Model} & \textbf{NitiBench Accuracy}  \\
\hline
GRPO & 15.82\% \\
\midrule
\textbf{InK-GRPO PT} & \textbf{19.30\%} \\
\textbf{InK-GRPO SFT} & 16.89\% \\
\hline
\end{tabular}
\caption{
NitiBench accuracy comparing GRPO with InK-GRPO variants that jointly optimize in-domain text using either pretraining-style corpora (PT) or SFT-style instruction data (SFT). Both variants improve over GRPO; however, the PT variant achieves the highest accuracy, suggesting that pretraining-style in-domain optimization more effectively complements RFT.
}
\label{tab:rl_q2}
\end{table}

\subsubsection{A7: Agentic RFT Further Improves Performance, and InK-GRPO Benefits Agentic Settings}

\Cref{tab:rl_q4} shows that \emph{InK-GRPO outperforms the GRPO baseline} in an agentic RFT setup, where the model operates in a multi-turn agentic scenario. Notably, this approach enables a 4B-parameter model to achieve task-specific performance exceeding that of GPT-5–level baselines. We further observe that, despite the availability of externalized knowledge through tool use (e.g., search), incorporating in-domain pretraining-style data as in InK-GRPO yields additional gains over a model trained with GRPO alone. We hypothesize that these improvements arise because in-domain corpora facilitate the acquisition of atomic task-relevant skills, enabling the model to more effectively learn complex behaviors necessary for difficult tasks during RFT \citep{yuan2025fxgxfgxllms}.

\begin{table}[ht]
\centering
\small
\renewcommand*{\arraystretch}{1.2}
\begin{tabular}{l|c}
\hline
\textbf{Model} & \textbf{NitiBench Accuracy} \\
\hline
Qwen3-4B-Instruct-2507 + Agent & 46.11\% \\
GPT-5 + Search & 38.07\% \\
GPT-5 + Agent & 75.34\% \\
\midrule
Agentic GRPO & 73.73\% \\
\midrule
\textbf{Agentic InK-GRPO} & \textbf{78.02\%} \\
\hline
\end{tabular}
\caption{Agentic (multi-turn) evaluation on NitiBench: InK-GRPO with agentic RFT outperforms GRPO under the same agentic retrieval setup and achieves higher accuracy than GPT-5 + Agent, which was evaluated with comparable tool-augmented configurations, and much higher than GPT-5 + Search, which is GPT-5 with built-in search.}
\label{tab:rl_q4}
\end{table}

\subsubsection{A8: InK-GRPO Preserves General Capabilities Without Evidence of Severe Catastrophic Forgetting}
To assess general capability retention and potential catastrophic forgetting, we evaluate all post-trained models from RQ5–RQ7 using the same general evaluation suite described in \Cref{sec:base2inst_eval}. This suite spans English and Thai chat, instruction following, knowledge, math, code, and tool/agent benchmarks.

Overall, \Cref{tab:rl_q5} shows that performance remains stable: across all settings, average scores fall within a narrow range (48.08\%-49.55\%) relative to the base \textbf{Qwen3 Instruct} baseline (48.07\%), regardless of the training algorithm. We emphasize that these models are trained exclusively on either NitiBench or MIRAGE-Bench, rather than on general-purpose benchmarks. These results indicate that neither GRPO nor InK-GRPO induces broad degradation of general capabilities, and in several cases both methods yield net improvements.

We observe that performance changes are largely \emph{targeted} rather than global. For example, GRPO-NitiBench achieves the highest overall average (49.55\%) and improves chat and code-switching metrics without degrading performance on math or knowledge benchmarks. InK-GRPO variants typically closely track their GRPO counterparts, suggesting that the additional in-domain next-token objective does not harm generalization. This trend aligns with prior observations that RFT methods generally preserve existing knowledge and exhibit limited catastrophic forgetting \citep{chu2025sft,zhu2025the,chen2025retainingdoingroleonpolicy,shenfeld2025rls}.

\begin{table}[!ht]
\tabcolsep=1mm
\centering
\renewcommand*{\arraystretch}{1.2}
\resizebox{\linewidth}{!}{
\begin{tabular}{l|cc|c|cc|ccc|cc|c|ccc|c}
\toprule
    \multirow{3}{*}{\textbf{Model}} &
    \multicolumn{2}{c|}{\textbf{Chat}} &
    \multicolumn{1}{c|}{\textbf{CS}} &
    \multicolumn{2}{c|}{\textbf{IF}} &
    \multicolumn{3}{c|}{\textbf{Knowledge}} &
    \multicolumn{2}{c|}{\textbf{Math}} &
    \multicolumn{1}{c|}{\textbf{Code}} &
    \multicolumn{3}{c|}{\textbf{Tool / Agent}} &
    \multicolumn{1}{c}{\multirow{3}{*}{\textbf{Average}}} \\
    & EN & TH & TH & EN & TH & EN & \multicolumn{2}{c|}{TH} & EN & TH & EN & EN & TH & EN &  \\
    & MTB & MTB & CS & IFE & IFE & GPQA & MMLU & OTE & M500 & M500 & LCB & HPQA & HPQA & BFCL & \\
    \midrule
    Qwen3 4B Instruct 2507 & 8.65 & \textbf{7.20} & 96.20 & 87.42 & 79.68 & 39.39 & 34.18 & 67.23 & \textbf{87.20} & 78.76 & 33.50 & 14.00 & 8.00 & \textbf{31.53} & 48.07 \\
    \midrule
    GRPO-MIRAGE & 8.56 & 7.04 & 95.00 & \textbf{88.03} & 79.44 & 43.43 & 33.50 & 67.40 & 87.00 & 79.76 & 39.71 & 19.00 & 6.00 & 30.95 & 48.92 \\
    InK-GRPO-MIRAGE & 8.47 & 7.07 & 96.00 & 86.61 & 78.01 & 41.41 & 33.16 & 67.23 & 86.00 & 79.76 & 39.54 & 9.00 & 10.00 & 30.80 & 48.08 \\
    \midrule
    GRPO-NitiBench & \textbf{8.84} & 6.96 & \textbf{98.20} & 87.66 & 77.50 & 41.41 & 34.35 & \textbf{67.91} & 86.40 & 79.16 & 39.05 & 16.00 & \textbf{19.00} & 31.27 & \textbf{49.55} \\
    InK-GRPO-NitiBench (PT) & 8.66 & 7.05 & 95.60 & 87.36 & 78.37 & 37.88 & \textbf{36.05} & 66.98 & 84.80 & 79.16 & 38.40 & 16.00 & 17.00 & 30.66 & 48.86 \\
    InK-GRPO-NitiBench (SFT) & 8.72 & 7.07 & 95.80 & 88.09 & \textbf{80.50} & 42.42 & 35.20 & 67.32 & 84.40 & \textbf{80.36} & 36.27 & 16.00 & 10.00 & 30.89 & 48.79 \\
    \midrule
    GRPO + Agent & 8.63 & 7.04 & 94.20 & 87.57 & 79.08 & 40.40 & 33.84 & 67.06 & 83.80 & 77.76 & \textbf{46.08} & 9.00 & 3.00 & 31.29 & 47.77 \\
    InK-GRPO + Agent & 8.68 & 6.95 & 94.60 & 87.72 & 78.85 & \textbf{44.44} & 34.01 & 65.36 & 83.80 & 77.76 & 42.48 & \textbf{24.00} & 12.00 & 30.28 & 49.35 \\
\bottomrule
\end{tabular}
}
\caption{Evaluation for general capability retention and catastrophic forgetting. Higher is better. Across settings, GRPO and InK-GRPO preserve general performance with no evidence of severe catastrophic forgetting.}
\label{tab:rl_q5}
\end{table}

\section{Summary: Typhoon-S Recipe}
\label{sec:recipe}

This section summarizes the final \textbf{Typhoon-S} post-training recipes for:
(1) \textbf{transforming a base model to an instruct model} for improved adoptability and
(2) building \textbf{a legal RAG agent} to support sovereign capability.
Both recipes are intentionally minimal, using only a few training stages, openly available datasets, transparent training details, and academic-scale compute. These recipes result in the following open releases of the models as: \textbf{Typhoon-S-8B Instruct}\footnote{\url{https://huggingface.co/typhoon-ai/typhoon-s-thaillm-8b-instruct-research-preview}} and \textbf{Typhoon-S-4B Legal Agent}\footnote{\url{https://huggingface.co/typhoon-ai/typhoon-s-4b-nitibench-ccl-legal-agent-research-preview}}.

\subsection{Typhoon-S-8B Instruct (Adoptability)}
\label{sec:recipe_8b}

We aim to develop a recipe for transforming a sovereignty-adapted \emph{base} model into a strong general-purpose \emph{instruction-tuned} assistant while preserving Thai-native strengths.
After validating the SFT+OPD recipe on Qwen3-4B in RQ1-RQ4 (\Cref{sec:adopt}), we apply the same approach to a sovereignty-adapted base model to obtain Typhoon-S-8B Instruct.

\subsubsection{Recipe}
We start from the sovereignty-adapted base model \texttt{ThaiLLM/ThaiLLM-8B}, an open model obtained by continued pretraining of \texttt{Qwen3-8B-Base} on 64B tokens of Thai corpus.
We then apply a two-stage post-training recipe--\textbf{SFT} followed by \textbf{OPD}--using \texttt{Qwen/Qwen3-30B-A3B-Instruct-2507} as the OPD teacher, which provides token-level distillation targets.

\paragraph{Data}
We use open English instruction data as the foundation and add a small, targeted Thai dataset for Thai alignment.
The SFT data mixture is shown in \Cref{tab:sft_data}, and the OPD mixture is shown in \Cref{tab:distill_data}.
In line with RQ3, Thai data is included at both stages to improve Thai-native performance.
The full training datasets for each stage are openly available on \href{https://huggingface.co/datasets/typhoon-ai/typhoon-s-instruct-post-training}{Hugging Face}.

\paragraph{Training Stages}
\begin{enumerate}
    \item \textbf{Stage 1: SFT}
    Full fine-tuning on a mixed instruction corpus (general instructions, tool use, and Thai alignment),
    optimizing cross-entropy loss on instruction-response pairs.

    \item \textbf{Stage 2: OPD}
    Full-logits OPD on student-generated trajectories and sequences sampled from the SFT corpus,
    minimizing the forward KL divergence (teacher-to-student) at the token level.
\end{enumerate}

\paragraph{Hyperparameters}
We use the hyperparameters listed in \Cref{tab:hyperparameters_sft_opd}.
In brief, SFT uses AdamW optimizer with learning rate of $2\times10^{-5}$ for 2 epochs with sequence packing up to 16,384 tokens.
OPD uses AdamW optimizer with learning rate of $1\times10^{-6}$ for 1 epoch with student-data fraction $\lambda=0.25$.
To make full-logits OPD feasible under limited GPU memory, we employ dynamic model swapping, FSDP with CPU offloading, and vLLM-backed rollouts.

\subsubsection{Result}
The resulting model, \textbf{Typhoon-S-8B Instruct}, improves Thai-native chat quality, Thai code-switching robustness, Thai knowledge (OTE), and Thai agentic retrieval QA compared to a strong multilingual instruction-tuned baseline (Qwen3-8B), while remaining competitive on general English benchmarks Detailed performance results are reported in\Cref{tab:sft_q51,tab:sft_q52}.

\subsection{Typhoon-S-4B Legal Agent (Sovereign Capability)}
\label{sec:recipe_4b_agent}

We aim to improve \textbf{Thai legal reasoning performance} in a sovereignty-focused setting by training a RAG agent capable of multi-turn tool use to retrieve and reason over an in-domain legal corpus.

\subsubsection{Recipe}
We start from \texttt{Qwen3-4B-Instruct-2507} as a general-purpose instruction-tuned backbone and adapt it to the Thai legal RAG setting via a single agentic RFT stage. Specifically, we apply the recipe developed in RQ5-RQ8, namely the \textbf{Agentic InK-GRPO} algorithm \Cref{sec:frontier}.

\paragraph{Data}
Training is performed on the \texttt{ccl} split of NitiBench.\footnote{\url{https://huggingface.co/datasets/airesearch/WangchanX-Legal-ThaiCCL-RAG}}
The \texttt{question} field provides the agent prompt, the \texttt{positive\_answer} field serves as the reference answer for reward computation, and the \texttt{positive\_contexts} field provides in-domain legal content used as pretraining-style data for CE loss computation by InK-GRPO. 
The training dataset is available on \href{https://huggingface.co/datasets/typhoon-ai/typhoon-s-sovereign-capability-dataset}{Hugging Face}.

\paragraph{Environment and Tools}
Training and evaluation are conducted in a controlled RAG environment with two tools:
\begin{itemize}[leftmargin=1.2em,itemsep=2pt,topsep=2pt]
    \item \texttt{search}: semantic retrieval over an in-domain legal corpus;
    \item \texttt{read}: returns the full content of a selected document.
\end{itemize}
Documents are embedded using \texttt{Qwen/Qwen3-Embedding-0.6B} and indexed in FAISS using an IVF-SQ8 configuration.
Each \texttt{search} call returns the top-3 documents, and \texttt{read} returns the full document content.
Tool outputs are masked from gradient computation during RFT.

\paragraph{Agentic InK-GRPO}
Typhoon-S-4B Legal Agent is trained using a single-stage \textbf{Agentic InK-GRPO} algorithm that combines:
\begin{enumerate}[leftmargin=1.2em,itemsep=2pt,topsep=2pt]
    \item \textbf{Trajectory-level Agentic GRPO.}
    The model interacts with the RAG environment over multiple tool-use turns, and GRPO is applied over entire trajectories using an accuracy-based reward.
    \item \textbf{Stochastic in-domain next-token prediction.}
    At each optimization step, an auxiliary cross-entropy loss on in-domain legal text is activated with probability $\rho$, injecting legal-domain knowledge during RFT.
\end{enumerate}
Formally, the training loss at each step is
\begin{equation}
\mathcal{L} = \mathcal{L}_{\text{GRPO}} + \lambda\,b\,\mathcal{L}_{\text{CE}}, \quad b \sim \text{Bernoulli}(\rho),
\end{equation}
where $\mathcal{L}_{\text{GRPO}}$ is computed from on-policy agent trajectories, and $\mathcal{L}_{\text{CE}}$ is computed on a separate in-domain, pretraining-style dataset.

\paragraph{Reward Function}
Agentic training uses an \textbf{accuracy-only} reward based on an LLM-as-a-judge comparison between the model's final answer and the reference answer.
The judge model operates on full generated trajectories but the reward is assigned solely based on the final answer; no additional tool-specific rewards are used.

\paragraph{Hyperparameters}
We use the RFT hyperparameters listed in \Cref{tab:hyperparameters_rl}.
Key settings include: AdamW optimizer with a learning rate of $1\times10^{-6}$, a sampling temperature of 0.7, DAPO-style decoupled clipping (clip-high 0.24, clip-low 0.20), overlong reward shaping, and InK-GRPO mixing with $\rho=0.6$ and $\lambda=0.1$.
Training is performed on 4$\times$H100 GPUs.

\subsection{Result}
The resulting model, \textbf{Typhoon-S-4B Legal Agent}, is trained with single-stage Agentic InK-GRPO and achieves the strongest NitiBench agentic accuracy among our 4B variants (\Cref{tab:rl_q4}), surpassing both GRPO-only training and GPT-5 with comparable agentic setups, while preserving general capabilities without evidence of severe catastrophic forgetting (\Cref{tab:rl_q5}).

\section{Conclusion}

We present \textbf{Typhoon S}, a minimal and open post-training recipe for sovereign LLMs that enables both \emph{adoptability} and \emph{sovereign capability} under limited resources.
Using Thai as a case study, we show that SFT and OPD are sufficient to transform \emph{base} models into competitive \emph{instruction-tuned} assistants, while small-scale RFT with \textbf{InK-GRPO} substantially improves Thai legal reasoning and agentic retrieval without degrading general capabilities. These results are achieved with academic-scale compute, demonstrating that careful post-training design--not scale alone--can support practical, transparent, and sovereign LLMs.

\section*{Limitations \& Future Work}
While we extensively experiment with modern post-training approaches such as SFT, OPD, and RFT in this report, we do not explore pre-training or mid-training, as these experiments require resources at a larger scale than are currently available to us. Future studies can extend our findings by investigating the dynamics of pre- and mid-training under our proposed approaches. Our study focuses on Thai because we have access to the necessary data and domain expertise and, as language owners, possess the cultural and linguistic knowledge required for effective evaluation. However, we believe that our proposed approaches and findings generalize to other settings and are not specific to Thai (i.e., they are not tied to a fixed Thai dataset). Accordingly, we aim to examine how well the approach generalizes to other languages and cultural contexts in future work.

All experiments were conducted using 4$\times$H100 GPUs, except for the final run of section 2 (\texttt{Typhoon-S-8B-Instruct}), which was trained on 8$\times$H100 GPUs at a scale comparable to those used by many national and academic laboratories. Scaling to larger setups, or systematically studying the scaling behavior of these techniques, remains an important direction for future work.

\section*{Acknowledgments}
Beyond the primary authors, we gratefully acknowledge the Typhoon Team members at SCB 10X. We also extend our appreciation to the SCBx R\&D Team for their support, resources, and valuable insights. Lastly, we are grateful to the global and local AI communities for open-sourcing resources and sharing knowledge.

\bibliographystyle{plainnat}
\bibliography{refs}

\clearpage
\beginappendix

\section{Hyperparameters}

This section summarizes the hyperparameter configurations used in our post-training experiments. Unless otherwise noted, hyperparameters were selected based on preliminary tuning and held fixed across benchmarks to ensure comparability of results.

\subsection{Hyperparameter for SFT \& OPD}

\Cref{tab:hyperparameters_sft_opd} reports the hyperparameters used for SFT and OPD. These settings are shared across all benchmarks unless explicitly stated.

\begin{table}[!ht]
\centering
\small
\renewcommand*{\arraystretch}{1.2}
\begin{tabular}{lcc}
\toprule
\textbf{Hyperparameter} & \textbf{SFT} & \textbf{OPD} \\
\midrule
\rowcolor{typhoonpurple!20}\multicolumn{3}{l}{\textit{\textbf{Training Configuration}}}\\
Optimizer & AdamW & AdamW \\
Learning rate & $2 \times 10^{-5}$ & $1 \times 10^{-6}$ \\
Epochs & 2 & 1 \\
Global batch size & 32 & 64 \\
LR scheduler & Cosine & Cosine \\
Warmup ratio & 0.05 & 0.05 \\
Max sequence length & 16384 & 4096 \\
\midrule
\rowcolor{typhoonpurple!20}\multicolumn{3}{l}{\textit{\textbf{OPD Specific}}}\\
$\lambda$ (student-data fraction) & -- & 0.25 \\
Sampling temperature & -- & 1.0 \\
Top-$p$ & -- & 0.95 \\
Max completion length & -- & 2048 \\
\bottomrule
\end{tabular}
\caption{Hyperparameters used for SFT and OPD experiments. In RQ 2, we specifically apply top-$k=10$ distillation, while other steps utilize the full distribution.}
\label{tab:hyperparameters_sft_opd}
\end{table}

\subsection{Hyperparameter for RFT}

\Cref{tab:hyperparameters_rl} lists the hyperparameters used for RFT experiments based on GRPO. In addition to standard GRPO settings, we include parameters for the stochastic auxiliary CE objective used in InK-GRPO, as well as agentic-specific configurations where applicable.

\begin{table}[htbp]
\centering
\small
\renewcommand*{\arraystretch}{1.2}
\begin{tabular}{lccc}
\toprule
\textbf{Hyperparameter} & \textbf{NitiBench} & \textbf{NitiBench-Agentic} & \textbf{MIRAGE-Bench} \\
\midrule
\rowcolor{typhoonpurple!20}\multicolumn{4}{l}{\textit{\textbf{GRPO}}}\\
Optimizer & AdamW & AdamW & AdamW \\
Learning rate & $1 \times 10^{-6}$ & $1 \times 10^{-6}$ & $1 \times 10^{-6}$ \\
Epochs & 2 & 1 & 3 \\
GRPO epoch per step & 1 & 1 & 1 \\
Optimization steps & 512 & 256 & 204 \\
Global batch size & 32 & 32 & 32 \\
Mini-batch size & 32 & 8 & 32 \\
KL with ref policy & No & No & No \\
LR scheduler & Constant & Constant & Constant \\
Sampling temperature & 0.7 & 0.7 & 0.7 \\
Max response length & 4096 tokens & 8192 tokens & 4096 tokens \\
Clip-high ratio & 0.24 & 0.24 & 0.24 \\
Clip-low ratio & 0.20 & 0.20 & 0.20 \\
Overlong reward shaping & \checkmark & \checkmark & \checkmark \\
\midrule
\rowcolor{typhoonpurple!20}\multicolumn{4}{l}{\textit{\textbf{CE}}}\\
$\rho$ & 0.6 & 0.6 & 0.6 \\
$\lambda$ & 0.1 & 0.1 & 0.1 \\
CE batch size & $64$ & $16$ & $64$ \\
CE update frequency & Per mini-batch & Per mini-batch & Per mini-batch \\
\midrule
\rowcolor{typhoonpurple!20}\multicolumn{4}{l}{\textit{\textbf{Agentic}}}\\
Max user turns & -- & 5 & -- \\
Max tool response length & -- & 1024 tokens & -- \\
\bottomrule
\end{tabular}
\caption{Hyperparameters used for RFT experiments across NitiBench, NitiBench-Agentic, and MIRAGE-Bench. GRPO denotes standard RFT settings; CE denotes the stochastic auxiliary cross-entropy objective used in InK-GRPO. Agentic-specific hyperparameters are applied only to NitiBench-Agentic.}
\label{tab:hyperparameters_rl}
\end{table}

\section{Prompt Templates}

This section provides the complete prompt templates used throughout our experiments. These prompts define how we evaluate model responses and compute reward signals during training.

\subsection{RL Reward Calculation Prompt}

We use an LLM-as-a-judge approach to compute rewards during RFT. The prompt in \Cref{prompt:training_judge} instructs a judge model to evaluate responses on a 3-point scale (0-2) based on task-specific criteria. The judge assesses whether responses meet all specified requirements and returns structured feedback including both a numerical score and explanation.

\begin{figure}[!ht]
\centering
\footnotesize
\begin{tcolorbox}
You are an expert evaluator. Your task is to evaluate how well a response addresses a given prompt according to specific evaluation criteria.\\
\\
\# Task\\
Evaluate the response below using a 3-level scoring system:\\
- **Score 0**: The response is incorrect, irrelevant, or does not address the requirements\\
- **Score 1**: The response partially addresses the requirements but has significant gaps, errors, or missing information\\
- **Score 2**: The response fully addresses all requirements correctly and completely\\
\\
\# Evaluation Criteria\\
\{llm\_judge\_prompt\}\\
\\
\# User Prompt\\
\{prompt\}\\
\\
\# Response to Evaluate\\
\{response\_to\_judge\}\\
\\
\# Instructions\\
1. Carefully check if the response meets ALL requirements specified in the evaluation criteria\\
2. Assign a score of 0, 1, or 2 based on how well it meets the criteria\\
3. Provide a brief explanation justifying your score\\
4. Return your evaluation in the following JSON format:\\
\\
\{\{\\
\quad\textquotesingle\textquotesingle score\textquotesingle\textquotesingle : <0, 1, or 2>,\\
\quad\textquotesingle\textquotesingle explanation\textquotesingle\textquotesingle : \quad\textquotesingle\textquotesingle <Brief explanation of why you gave this score>\textquotesingle\textquotesingle \\
\}\}\\
\\
\texttt{\{ground truth\}}
\end{tcolorbox}
\caption{Prompt template for calculating RFT rewards using LLM-as-a-judge evaluation.}
\label{prompt:training_judge}
\end{figure}

\subsection{LLM Judge Prompt}

We evaluate model responses using an LLM-based judge that determines correctness by comparing a generated answer against a reference solution. The judge is instructed to assess whether the model output matches the ground-truth answer exactly, without partial credit or subjective interpretation. \Cref{prompt:evaluate_judge} shows the prompt template used for this evaluation.

\begin{figure}[!ht]
\centering
\footnotesize
\begin{tcolorbox}
To correctly answer this question, the response must match the following answer:\\
\\
\texttt{\{ground\_truth\}}
\end{tcolorbox}
\caption{LLM-judge prompt used to compute an accuracy score by comparing model responses against a ground-truth answer.}
\label{prompt:evaluate_judge}
\end{figure}

\section{Pseudo-code For InK-GRPO}

\Cref{appendix:rl_pretrain_code} details the InK-GRPO training procedure, which combines on-policy GRPO with stochastic in-domain language modeling (CE objective). At each training step, the algorithm: (1) collects on-policy rollouts and computes the GRPO objective, and (2) stochastically mixes in a cross-entropy loss on in-domain text data with probability $\rho$ and weight $\lambda$. This approach optimizes task-specific rewards while teaching knowledge.

\begin{algorithm}[t]
\caption{InK-GRPO Training (On-Policy GRPO with Stochastic In-Domain CE Augmentation)}
\label{appendix:rl_pretrain_code}
\begin{algorithmic}[1]
\Require Policy model $\pi_\theta$; optional reference policy $\pi_{\text{ref}}$;
reward function $R(\cdot)$; prompt distribution $\mathcal{D}_{\text{prompt}}$;
in-domain text corpus $\mathcal{D}_{\text{CE}}$;
mixing probability $\rho$; CE weight $\lambda$;
rollouts per prompt $K$; number of training steps $T$;
GRPO epochs per step $E$; GRPO minibatches per step $M$; learning rate $\eta$
\For{$t = 1$ \textbf{to} $T$}
  \State \textit{// On-policy GRPO data collection}
  \State Sample prompts $\mathcal{P} \sim \mathcal{D}_{\text{prompt}}$
  \State For each $p \in \mathcal{P}$, sample $K$ responses: $\{y^{(k)}\}_{k=1}^K \sim \pi_\theta(\cdot \mid p)$
  \State Compute rewards: $r^{(k)} \gets R(p, y^{(k)})$ for all $p \in \mathcal{P}, k \in [K]$
  \State Construct GRPO training buffer $\mathcal{D}_{\text{GRPO}} \gets \{(p, y^{(k)}, r^{(k)})\}$
  \Statex

  \State \textit{// Optimize on collected data with minibatches}
  \For{$e = 1$ \textbf{to} $E$}
    \State Shuffle $\mathcal{D}_{\text{GRPO}}$
    \For{$m = 1$ \textbf{to} $M$}
      \State Sample GRPO minibatch $\mathcal{B}_{\text{GRPO}} \sim \mathcal{D}_{\text{GRPO}}$
      \If{$\pi_{\text{ref}}$ is provided}
        \State $\mathcal{L}_{\text{GRPO}} \gets \textsc{GrpoObjective}(\pi_\theta, \mathcal{B}_{\text{GRPO}}, \pi_{\text{ref}})$
      \Else
        \State $\mathcal{L}_{\text{GRPO}} \gets \textsc{GrpoObjective}(\pi_\theta, \mathcal{B}_{\text{GRPO}})$
      \EndIf

      \Statex
      \State \textit{// Stochastic in-domain CE augmentation (per minibatch update)}
      \State Sample $b \sim \text{Bernoulli}(\rho)$
      \If{$b = 1$}
        \State Sample in-domain batch $\mathcal{B}_{\text{CE}} \sim \mathcal{D}_{\text{CE}}$
        \State $\mathcal{L}_{\text{CE}} \gets \textsc{NextTokenLogLik}(\pi_\theta, \mathcal{B}_{\text{CE}})$
        \State $\mathcal{L} \gets \mathcal{L}_{\text{GRPO}} + \lambda \, \mathcal{L}_{\text{CE}}$
      \Else
        \State $\mathcal{L} \gets \mathcal{L}_{\text{GRPO}}$
      \EndIf

      \Statex
      \State \textit{// Update parameters}
      \State $\theta \gets \theta + \eta \, \nabla_\theta \mathcal{L}$
    \EndFor
  \EndFor
\EndFor
\end{algorithmic}
\end{algorithm}

\end{document}